\documentclass{article}
\usepackage{ijcai20}

\usepackage{times}
\usepackage{soul}
\usepackage{url}
\usepackage[hidelinks]{hyperref}
\usepackage[utf8]{inputenc}
\usepackage[small]{caption}
\usepackage{graphicx}
\usepackage{amsmath}
\usepackage{amsthm}
\usepackage{booktabs}
\usepackage{algorithm}
\usepackage{algorithmic}
\title{A decentralized aggregation mechanism for training deep learning models using smart contract system for bank loan prediction}

\author{
Pratik Ratadiya$^1$
\and
Khushi Asawa$^1$
\and
Omkar Nikhal$^1$
\affiliations
$^1$Dept. of Computer Engineering,\\
Pune Institute of Computer Technology,\\
Maharashtra, India.\\
\emails
prratadiya@gmail.com,
khushiasawa31@gmail.com,
nikhalomkar@gmail.com
}

\begin{document}

\maketitle

\begin{abstract}
  Data privacy and sharing has always been a critical issue when trying to build complex deep learning-based systems to model data. Facilitation of a decentralized approach that could take benefit from data across multiple nodes while not needing to merge their data contents physically has been an area of active research. In this paper, we present a solution to benefit from a distributed data setup in the case of training deep learning architectures by making use of a smart contract system. Specifically, we propose a mechanism that aggregates together the intermediate representations obtained from local ANN models over a blockchain. Training of local models takes place on their respective data. The intermediate representations derived from them, when combined and trained together on the host node, helps to get a more accurate system. While federated learning primarily deals with the same features of data where the number of samples being distributed on multiple nodes, here we are dealing with the same number of samples but with their features being distributed on multiple nodes. We consider the task of bank loan prediction wherein the personal details of an individual and their bank-specific details may not be available at the same place. Our aggregation mechanism helps to train a model on such existing distributed data without having to share and concatenate together the actual data values. The obtained performance, which is better than that of individual nodes, and is at par with that of a centralized data setup makes a strong case for extending our technique across other architectures and tasks. The solution finds its application in organizations that want to train deep learning models on vertically partitioned data.
\end{abstract}

\section{Introduction}
\par Artificial Intelligence (AI) and blockchain are proving to be one of the most powerful and dynamic pair of technologies used together, improving the performance of the other one in every discipline where they are implemented. They have been used together to cause major upgrades in various applications including supply chain logistics, finance-related data, and medical health records \cite{GeeksForGeeks}. Blockchains are decentralized ledgers which maintain an immutable record of transactional data in digital format \cite{zheng2018blockchain}. Deep learning is a sub-domain in AI that involves multi-layered architectures that extract features from the input data and map them to the output labels. Recent years have seen a rise in the widespread application of AI models and algorithms in various areas \cite{wang2018interactive,ratadiya2019attention}. Deep learning-based AI systems strongly rely on data, something which blockchains can store with an exceptionally high degree of authenticity. The Blockchain-AI convergence is imminent since both of them provide a distinct, significant advantage to entities which store data.
\par With the advancement of AI, many of the data privacy challenges have also been cited, especially when working on real-world tasks with the existing data \cite{jagannathan2005privacy,mohassel2017secureml}. Being able to re-identify personal information using large datasets, lack of transparency in the use of consumer data \cite{luminovo}, regulations by countries, and unions have been some major factors in intensifying this problem. With more features and personal data being collected and stored, there is an inherent risk and higher probability of sensitive information getting exposed and shared.
\par Federated learning has been an approach that has been used in the past to tackle data-sharing issues \cite{konevcny2016federated}. However, federated learning has mostly stressed on combining updates from models trained on different samples of data, but those which have similar types of features i.e. horizontally partitioned data \cite{smith2017federated}. However, many organizations also follow a vertical partitioning of data. Further, there might be hosts who wish to train a smart system based on data that is currently present at different organizations. The involved parties may not always be permitted to share the data. Consider the task of bank loan prediction. The best prediction model can be trained from the existing data by combining both the personal details of the users and also their bank or finance-related information. However, if these details are not present in a single place, quality feature extraction from such decentralized data is a challenging task.
\par In this paper, we propose a smart contract-based system for aggregation of features for better training from decentralized data. We prove the effectiveness of our architecture when working with artificial neural networks to build a model for the task of bank loan prediction. Setting up the architecture over a blockchain also makes the exchange of intermediate information more secure and free from external threats. The proposed mechanism is found to be superior in handling the data privacy issue and also provides optimum results on a publicly available dataset. Our main contributions in this paper could be listed as:
\begin{enumerate}
    \item We have been able to achieve quality results despite training the host model from two different sets of features extracted from local models which are located on separate nodes, as opposed to the conventional centralized model setup in deep learning.
    \item We have made effective use of the Ethereum Blockchain and IPFS to store, retrieve, and exchange the intermediate feature vectors required for the training, and test data required for the inferences from the local models.
    \item The performance of the host model trained on aggregated features from decentralized data is at par with a centrally trained model, thus providing a better alternative that is more private and secure as implemented on a blockchain.
\end{enumerate}
\par The rest of the paper is structured as follows: Section 2 talks about the related work in this area while the dataset description is presented in section 3. The proposed methodology is explained in section 4. Our obtained results and analysis of the work are discussed in section 5 while the paper is concluded in section 6.

\section{Background work}
\par In the domain of data privacy, several vulnerabilities have been detected. These have been consistently explored and worked upon by researchers to reduce the possibility of violation of any privacy norms, either unethically or legally. 
\par  Some of the earlier works are based upon the principle of secure multi-party computation(SMC) which has been used for k-means clustering \cite{bunn2007secure}, \cite{jagannathan2005privacy}, SVM classification \cite{lindell2000privacy}, linear regression functions \cite{du2004privacy}, stochastic gradient descent method \cite{mohassel2017secureml}, association rule mining in vertically partitioned data \cite{vaidya2002privacy}, and Naive Bayes classification \cite{vaidya2008privacy}. These approaches eliminated the trade-off between data usability and data privacy and safeguarded the system against scooping on the data. Privacy-preserving machine learning using a neural network has also been worked upon quite actively. \cite{xie2014crypto}, \cite{bos2014private} proposed a method in encrypted domain which applies homomorphic encryption to activation functions with neural networks to ensure the privacy of the data. \cite{barni2011privacy} used a privacy-preserving automatic diagnosis system based on linear branching programs and neural networks. \cite{pathak2011privacy} presented a model that contains an asymmetric framework for the encryption of probabilities that uses public-key additively homomorphic cryptosystem. They went on to propose a privacy-preserving speaker identification based on the Gaussian Mixture Model-based protocol \cite{pathak2012privacy}. Recent years have seen an active rise in the work being done in the area of federated learning, which proposes a way for combining weights or learning of the client nodes and returning them with a global update after aggregation of those weights \cite{konevcny2016federated}. Federated learning is also being tried out for multitask and semistructured learning \cite{smith2017federated}. Hardy et al. \cite{hardy2017private} proposed a novel federated learning approach for vertically partitioned data using both entity resolution and homomorphic encryption. However, the scope of their data modeling algorithms was restricted to logistic regression and Taylor approximation, which are not always sufficient, especially when modeling complex real-world data.
\par The majority of these previous approaches have applied secure multi-party computation. However, SMC models are subjected to computational overhead and high communication cost between the participants. Homomorphic encryption-based models using neural networks are computationally expensive. Although differential privacy safeguards privacy against a wide range of privacy attacks, it generates too much noise which ultimately reduces the data utility when dealing with diverse data. Thus, a more balanced solution, capable of working with separate data in an efficient yet effective way is the need of the hour.

\section{Dataset description}
\par We make use of a publicly available dataset\footnote{https://www.kaggle.com/itsmesunil/bank-loan-modelling} for personal bank loan classification. Given their details, the task is to predict whether a person will opt for the bank loan or not. 
\par The dataset consists of thirteen structured input columns and one binary output label, personal loan (0:loan not taken, 1:loan taken). The irrelevant features are eliminated using correlation analysis of individual columns with the output variable. We introduce a combination\_feature based on income and average card spending. After these steps, we are left with 10 input features. The dataset consists of 5000 samples, 480 of which are having output label 1 (loan is taken). The dataset is split into training and testing sets in the ratio of 7:3.
\par For our demonstration of decentralized setup, we split the training dataset into two parts: personal details (6 features) and bank-specific details (4 features). The two parts consist of the following columns:
\begin{itemize}
    \item Personal details: Education level, no. of family members, annual income, average credit card spending, years of work experience, the value of house mortgage.
    \item Bank-specific details: Does the customer have a security account with the bank, do they have a CD account with the bank, do they have a credit card issued by the Universal Bank, combination\_feature. 
\end{itemize}
\par Each of these splits are now considered as separate data nodes as could be the case in real life. Our objective is to train a model that can capitalize on both these splits during training while still not needing to explicitly share these values, hence preserving the data privacy regulations.

\section{Proposed methodology}
\par We work on the task of building a deep learning model for bank loan prediction. The existing training data is partitioned and available on two different nodes: one containing the personal details, and others containing the bank-specific details of the users. This is achieved by loading the respective data on two distinct nodes of the Ethereum blockchain. The nodes perform training on their local data and send the intermediate representations to the host. The host combines these representations and trains a third model based on which the final predictions will be derived. We explain our mechanism by describing our approach for each phase in the data modeling pipeline- training local models on respective data nodes, passing feature representations of training data to the host node, model training on the host node, and data flow for a new test input at the host.

\subsection{Training the models on individual nodes:}
\par We train two models on the respective nodes based on their data. As the data is structured in nature, we make use of the feed-forward neural network (FFNN) \cite{svozil1997introduction} setup, consisting of 3 hidden layers. After iterative designs, we stick to the following FFNN set up in this paper: For N input nodes, we have N nodes in the first and third hidden layer and 2N nodes in the second hidden layer. The output layer consists of a single node indicating the binary nature of the output.
\par After training the models, we extract the feature representations for each of the training samples from the third hidden layer. This layer being the last hidden layer consists of the richest quality feature vectors, and these obtained feature representations for the entire training data are stored by us in a Portable Document Format, on both the nodes. 

\subsection{Passing the intermediate feature representations of training data to the host node:}
\par The two files generated in the previous phase are to be passed to the host node, where they will be concatenated. To pass these files, we make use of the Ethereum Blockchain and the IPFS. Here, Ethereum serves as the backend whereas IPFS serves as the front-end. The InterPlanetary File System or IPFS is a peer-to-peer network used for storage and exchange of data in a distributed setup. Any user in the network can serve a file by its content address, and other peers in the network can find and request that content from any node that has it using a distributed hash table. Since IPFS and Blockchains are similar in structures, they go well together.
\par There is one potential flaw here. As long as anyone has the hash of the file, they can retrieve it from IPFS. We tackle this problem by encrypting the file using the asymmetric encryption technique. Asymmetric encryption encrypts the file with the public key of the intended recipient so that only they can retrieve the file since this file can only be decrypted by their private key. There are many encryption software, among which we are using the GNU Privacy Guard (GPG) software. We use PDF as the file format as it is best supported for data retrieval in IPFS and also makes the encryption process easier.
\par
\begin{figure}[t]
    \centering
    \includegraphics[height=6cm]{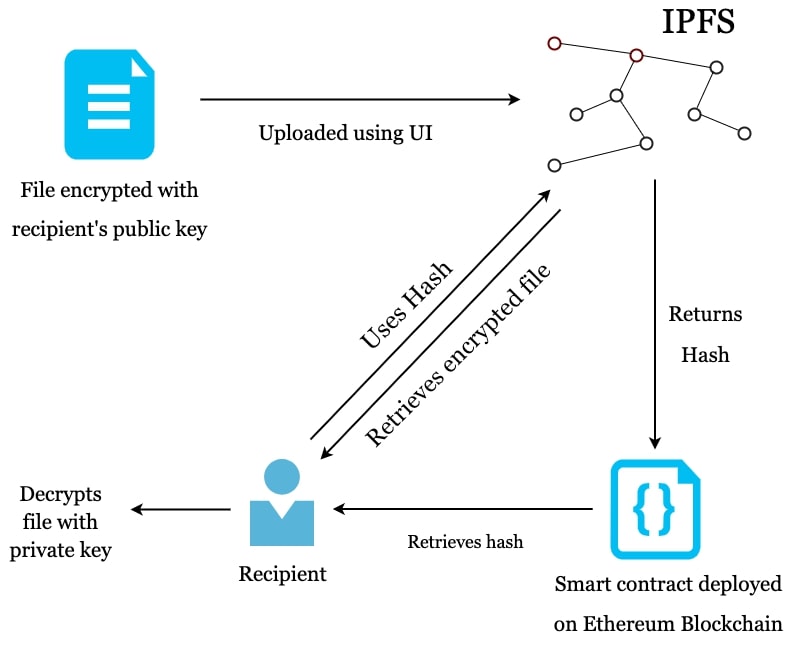}
    \caption{Flow of storing and retrieving the file using IPFS}
    \label{fig:ipfs}
\end{figure}

 The previously generated files containing the feature representations of the training data is encrypted with the host's public key and then are individually uploaded on the IPFS using a User-Interface. IPFS generates a hash of the file. This hash is then stored in an already deployed smart contract on the Ethereum Blockchain. The smart contract is a code that controls execution, and transactions are transparent, trackable, and irreversible. The host can now call a function of the smart contract to retrieve this hash. Using that hash, the host retrieves the file from IPFS and decrypt the file with the private key. By the preceding process, the host can retrieve both the files uploaded by the individual nodes. The nodes are expected to possess the host's public key. The sequence diagram of the flow of information is shown in Figure \ref{fig:ipfs}.

\begin{table}[H]
\renewcommand{\arraystretch}{1.2}
\begin{tabular}{llll}
\toprule
\textbf{Node} & \textbf{\begin{tabular}[c]{@{}l@{}}Nature of \\ input data\end{tabular}}         & \textbf{\begin{tabular}[c]{@{}l@{}}No. of \\ input nodes\end{tabular}}                   & \textbf{\begin{tabular}[c]{@{}l@{}}FFNN model\\ structure\end{tabular}} \\ \midrule
Node 1`       & Personal details                                                                 & 6                                                                                        & 6-6-12-6-1                                                        \\ 
Node 2        & \begin{tabular}[c]{@{}l@{}}Bank specific\\ details\end{tabular}                  & 4                                                                                        & 4-4-8-4-1                                                         \\ 
Host node     & \begin{tabular}[c]{@{}l@{}}Intermediate\\ feature\\ representations\end{tabular} & \begin{tabular}[c]{@{}l@{}}10\\ (6 dim+4 dim\\ vector from\\ node 1, node 2)\end{tabular} & 10-10-20-10-1                                                     \\ \bottomrule
\end{tabular}
\caption{Details of the deep learning models on each node. The second column indicates the type of values on which the respective model is trained, third column indicates number of input variables accepted by the respective model while the last column shows the number of nodes present in each layer of the respective model.}
\label{tabchar}
\end{table}

\subsection{Host node model training:}
\par
\begin{figure}[t]
    \centering
    \includegraphics[height=7cm]{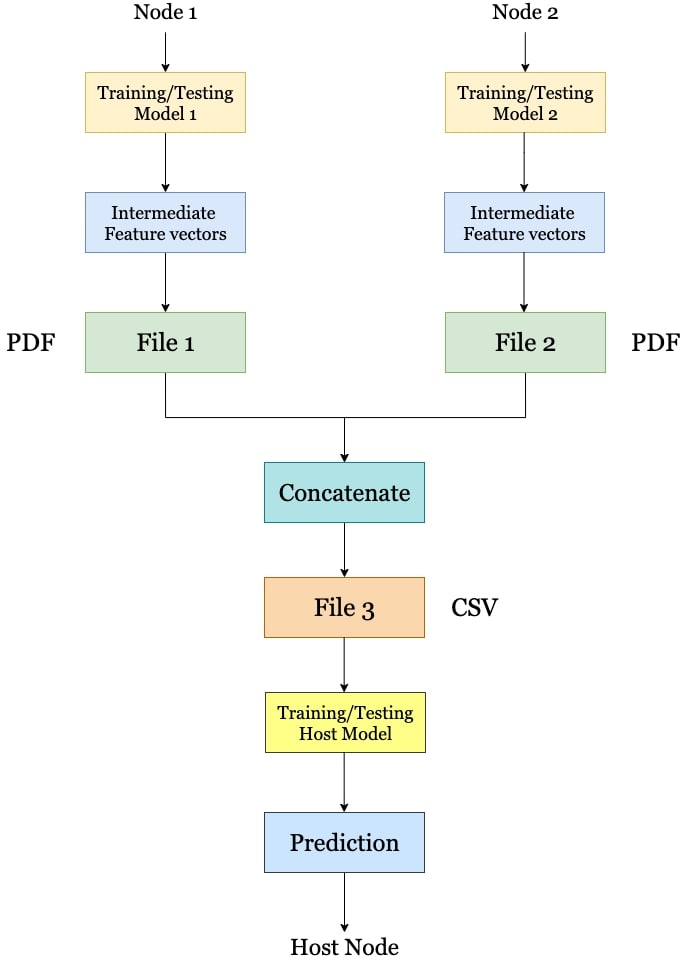}
    \caption{Block diagram of the proposed architecture(training)}
    \label{fig:train}
\end{figure}

 The host has the intermediate feature representations from both the nodes, which we concatenate together as a Comma-separated values (CSV) file. The concatenation or joining is done based on the common ID for each case. Now we train a model on the host where we map these feature representations to the output labels for the training data. The host's feed-forward neural network model has a similar architecture setup like that of the nodes, consisting of three hidden layers. The characteristics of the deep learning models for each of the three models are tabulated in Table \ref{tabchar}. The block diagram of the proposed architecture, concerning the training phase is shown in Figure \ref{fig:train}

\subsection{Prediction for a new test sample on the host node}
\par Now, say a new test case, of both details of the user, is passed to the host node. The output prediction is obtained with the following steps- transferring the specific columns from the host to the respective nodes, obtaining back the intermediate feature representation, passing this representation to the host model to derive the final prediction.

\subsubsection{Transferring the split of columns from the host to the respective nodes:}
\par A similar two split of the testing data instance is done and given to the respective nodes by the host as a portable document format. To pass the files, we make use of the Ethereum Blockchain and the IPFS in a similar way as we have used before. These files are encrypted with the public keys of the respective nodes and then are uploaded to the IPFS using a User-Interface. IPFS generates hashes of these files. These hashes are then stored in a smart contract on the Ethereum Blockchain. The nodes now call the functions of the smart contract to retrieve the hashes. Using the hashes, the nodes retrieve the testing dataset files from IPFS and decrypt it with their private keys. The host is supposed to be possessing the public keys of the two nodes.

\subsubsection{Obtaining the intermediate feature representation for the test data}
\par The two models present on the individual nodes are passed the received test data. We store the intermediate testing feature vectors, which are the output of the third hidden layer of the deep neural network in separate files of portable document format. The intermediate outputs are now passed back to the host.

\subsubsection{Passing the intermediate testing vectors to the host node:}
\par The two files generated in the last step are to be given to the host node using the Ethereum Blockchain and the IPFS in a similar way as before, where they will be concatenated. These files are encrypted with the host's public key and then are uploaded on the IPFS using a User-Interface. IPFS generates hashes of the file, which are stored in a smart contract on the Ethereum Blockchain. The host now retrieves the hashes from the smart contract function. Using these hashes, the host now retrieves the files from the IPFS and decrypts them using the private key. 

\subsubsection{Final prediction:}
\par The host has the intermediate testing feature vectors of both the files, which are concatenated together and passed to its model. The obtained probability from the softmax is rounded off to get the final prediction.
\par Each of the models is trained on the binary cross-entropy loss, with an Adam optimizer. The training takes place for 50 epochs. The hidden layers consist of the ReLU activation, while the output layers at all three models consist of the softmax activation function. The complete working of the system is shown in Algorithm 1.

\begin{algorithm}[h]
 \caption{Algorithm for Aggregation mechanism} 
 \begin{algorithmic}[1]
 \renewcommand{\algorithmicrequire}{\textbf{Input:}}
 \renewcommand{\algorithmicensure}{\textbf{Output:}}
 \REQUIRE Input data T ($T_1, T_2$), Output label O, Nodes N ($N_1, N_2$), Host node H
 \ENSURE  Prediction value P
 \STATE Initialize nodes N on the blockchain
 \STATE Load the respective training data on the nodes
 \FOR {each N nodes}
  \STATE Train the respective model $M_i$ on data
  \STATE Retrieve feature representation $F_i$ from third hidden layer and send to H
  \ENDFOR
  \STATE F = Concatenate($F_1, F_2$)
  \STATE Train model $M_H$ on H by mapping F to Y
  \FOR {new test case $T_{test}$}
  \STATE Send respective columns of $T_{test}$ to those nodes
  \STATE Obtain back the feature representations $F_{test}$ from them
  \STATE P = $M_H$.predict($T_{test}$)
  \ENDFOR
 \end{algorithmic}
 \end{algorithm}

\par We intend to show that the performance of $M_H$ is better than that of models $M_1$ and $M_2$.

\section{Results and analysis}
\par The results of the models are evaluated on their respective datasets as well as the combined datasets. We first compare the performance based on the accuracy obtained by the models on the respective nodes in Table \ref{compacc}.

\begin{table}[h]
\centering
\renewcommand{\arraystretch}{1.2}
\begin{tabular}{ll}
\toprule
\textbf{Model}                                                                                    & \textbf{Accuracy} \\
\midrule
N1: Personal details                                                                              & 97.8              \\ 
N2: Bank specific details                                                                         & 90.93             \\ 
\textbf{\begin{tabular}[c]{@{}l@{}}H: Aggregate trained model\\ (Proposed approach)\end{tabular}} & \textbf{98.13}    \\ 
\bottomrule
\end{tabular}
\caption{Comparison of accuracy obtained by the respective models}
\label{compacc}
\end{table}

\par It can be seen that our proposed aggregated mechanism has helped to come up with a host model better than the individual models trained on their respective data in terms of accuracy. To further solidify our claims, we also evaluate the performance in terms of their precision, recall, and F1 score and show our results in Table \ref{compall}.

\begin{table}[h]
\renewcommand{\arraystretch}{1.2}
\begin{tabular}{llll}
\toprule
\textbf{Model}                                                                                    & \textbf{Prec.} & \textbf{Recall} & \textbf{F1 score} \\
\midrule
N1: Personal details                                                                              & 0.96               & 0.92   & 0.94              \\ 
N2: Bank specific details                                                                         & 0.79               & 0.63            & 0.67              \\ 
\textbf{\begin{tabular}[c]{@{}l@{}}H: Aggregate trained\\ model (Proposed approach)\end{tabular}} & \textbf{0.97}      & \textbf{0.92}            & \textbf{0.95}     \\
\bottomrule
\end{tabular}
\caption{Comparison of respective models across multiple performance metrics}
\label{compall}
\end{table}

\par Thus it can be seen that our aggregated model has been able to combine the learned representations of both the models which have helped in an enhanced performance across all performance metrics. We have been able to train well from decentralized data setups and achieve quality results on new test cases.
\par It may seem obvious to few that combination and more data columns will give better results than involved sub-models. However, we focus on the aggregation within a decentralized environment and thus also compare our decentralized model performance with the third kind of model: the traditional centralized setup. We train and test a model directly on the 10 input values gathered together in one place in a conventional format. The classification reports for our decentralized model and the centralized model are tabulated in Table \ref{decentr} and \ref{centr} respectively.

\begin{table}[H]
\begin{tabular}{ccrrr}
\toprule
             & Precision            & Recall & F1-score & Support \\ \midrule
0            & 0.98                 & \textbf{1.00}   & 0.99     & 1343    \\
1            & \textbf{0.96}                 & 0.85   & 0.91     & 157     \\
Accuracy     & \multicolumn{1}{l}{} &        & 0.9813   & 1500    \\
Macro avg    & 0.97                 & 0.92   & 0.95     & 1500    \\
Weighted avg & 0.98                 & 0.98   & 0.98     & 1500    \\ \bottomrule
\end{tabular}
\caption{Classification report of model trained on decentralized data}
\label{decentr}
\end{table}

\begin{table}[H]
\begin{tabular}{ccrrr}
\toprule
             & Precision            & Recall & F1-score & Support \\ \midrule
0            & \textbf{0.99}                 & 0.99   & 0.99     & 1343    \\
1            & 0.95                 & \textbf{0.89}   & 0.92     & 157     \\
Accuracy     & \multicolumn{1}{l}{} &        & 0.9833   & 1500    \\
Macro avg    & 0.97                 & 0.94   & 0.95     & 1500    \\
Weighted avg & 0.98                 & 0.98   & 0.98     & 1500    \\ \bottomrule
\end{tabular}
\caption{Classification report of model trained on centralized data}
\label{centr}
\end{table}

\par It can be seen that there is not much difference in the performance of our decentralized model as compared to that of the centralized model across all the metrics. While the centralized model does have better recall and F1 score value for the minority class, it comes at the expense of having to share and combine two different data sources, thus introducing security and privacy concerns. There is only a 0.2\% difference between their accuracies, thus making a strong case for our decentralized setup. However, such a vertically partitioned dataset must not destroy the learning experience and performance of the model, and hence data partitioning needs to be appropriate.
\begin{figure}[h]
    \centering
    \includegraphics[height=5cm]{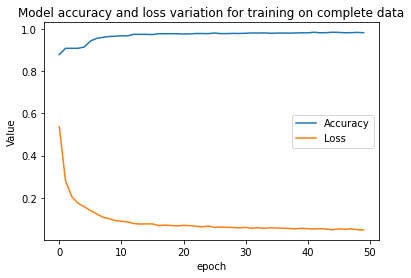}
    \caption{Training accuracy and loss progression for centralized data}
    \label{centrplt}
\end{figure}

\begin{figure}[h]
    \centering
    \includegraphics[height=5cm]{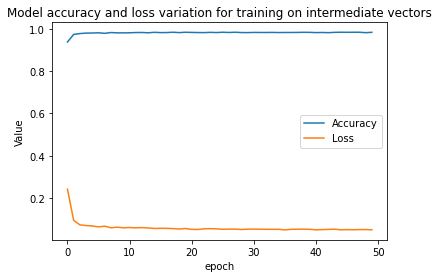}
    \caption{Training accuracy and loss progression of the host model for decentralized data}
    \label{decentrplt}
\end{figure}
\par Further, we also plot the training accuracies and loss variations to the epochs in the case of both the centralized and decentralized models. Their respective plots are visualized in Figure \ref{centrplt} and \ref{decentrplt}.
\par It can be seen that the decentralized model converges faster as compared to the centralized data model. This could be attributed to the fact that the decentralized data model receives richer data features and thus gets a better initialization, hence leading to faster convergence. We have thus been able to propose a system which gives at par results even with distributed, decentralized data and yet has a faster convergence period as compared to previous approaches.
\par Our proposed aggregation mechanism is more secure as it operates over a blockchain, makes use of an efficient file hashing mechanism, gives almost the same results on decentralized data as compared to centralized datasets, and hence makes for a strong case for organizations who wish to train models on their vertically partitioned data.

\section{Conclusion}
\par Despite having a decentralized setup, we have been able to accomplish standard results for the task of bank loan prediction using an effective aggregation mechanism. The performance of the aggregated model is better than the local nodes both in terms of performance metrics as well as convergence time. It is also almost equivalent to the performance of a centralized data setup. Blockchain and Artificial Intelligence can prove a powerful integrated architecture in preserving data privacy and maintaining data security. While the majority of the previous approaches have demonstrated their work on a centralized data format, our results show that using the aggregation of feature representations, we can still obtain maximum performance. Further improvements in our work include the use of more secure encryption techniques for encrypting the files before uploading them to IPFS. The mechanism can also be extended further for image and time series based data where more advanced deep learning models would be required. In the future, techniques that can extract information from multiple data sources without needing to share the actual data would be desired and will lay the foundation for a more secure, ethical, and private data analysis environment.

\bibliographystyle{named}
\bibliography{ijcai20}
\end{document}